\documentclass[final,times,twocolumn,authoryear]{elsarticle}

\makeatletter
\def\ps@pprintTitle{%
  \let\@oddhead\@empty
  \let\@evenhead\@empty
  \let\@oddfoot\@empty
  \let\@evenfoot\@empty
}
\makeatother
\usepackage{amssymb}
\usepackage{amsmath}
\usepackage{booktabs}
\usepackage{lmodern}
\usepackage{graphicx}
\usepackage{xcolor}
\usepackage{colortbl}
\usepackage{rotating}
\usepackage{makecell}
\usepackage{multirow}
\usepackage{hyperref}
\usepackage{natbib}
\usepackage{lineno}
\usepackage{float} 
\usepackage{pdflscape} 

\journal{}

\begin{document}

\newcommand{\citealiastext}[1]{\citetalias{#1}}

\begin{frontmatter}

\title{Task-Conditioned Synthetic Data Generation for Improving Machine Learning Performance in Agricultural Prediction Tasks}

\author{
Hamid Ebrahimy\textsuperscript{1,2,*}\,
Moritz Lucas\textsuperscript{1,2}\,
Martin Atzmueller\textsuperscript{1,3}\\[3pt]
\footnotesize \textsuperscript{1}Osnabrück University, Joint Lab Artificial Intelligence and Data Science, Osnabrück, Germany\\
\footnotesize \textsuperscript{2}Leibniz Institute for Agricultural Engineering and Bioeconomy (ATB), Potsdam, Germany\\
\footnotesize \textsuperscript{3}German Research Center for Artificial Intelligence (DFKI), Research Department Cooperative and Autonomous Systems (CAS), Osnabrück, Germany\\
\footnotesize \textsuperscript{*}Corresponding author: \texttt{hamid.ebrahimy@uni-osnabrueck.de}
}

\begin{abstract}
Machine Learning (ML) algorithms have been widely used to estimate agricultural variables across diverse contexts. However, because the quantity and quality of training data strongly influence performance of ML algorithms, their use can be constrained by limited or incomplete reference data. Synthetic Data Generation (SDG) offers a practical approach to address this issue by producing artificial but realistic samples that preserve key characteristics of the original data. Building on teacher–student knowledge transfer and in-context learning for tabular data, this study proposes a Task-Conditioned SDG (TCSDG) algorithm that pairs a Bayesian Network generator with a transformer-based tabular foundation model (TabICL). The proposed algorithm was evaluated on two agricultural prediction tasks: crop yield prediction and crop type classification. Six benchmark SDG algorithms were also utilized to compare their performance with that of TCSDG. Across twelve study sites, two training-data fractions, four multiplication ratios, and three predictive ML algorithms, augmenting the original data with TCSDG-generated synthetic data improved ML performance in 89\% of the crop type classification experiments and 74\% of the crop yield prediction experiments. TCSDG also substantially outperformed benchmark SDG algorithms and was the only method to consistently improve ML performance across both tasks at the aggregate level. The study demonstrates that carefully designed and processed synthetic data can improve ML performance in precision-agriculture applications. TCSDG offers a practical and extensible framework for generating synthetic data that supports downstream ML agricultural prediction. The full implementation of TCSDG is publicly available as open source at \href{https://github.com/HamidEbrahimy/TCSDG}{https://github.com/HamidEbrahimy/TCSDG}. 

\end{abstract}

\end{frontmatter}

\section{Introduction}

Agricultural variables are critical indicators that have been used in various applications including food security assessment, agricultural optimization, economic planning, and climate change impact evaluation \citep{bouguettaya_deep_2022,benedetti_accuracy_2010, van_klompenburg_crop_2020}. With the population growth and the increasing impact of climate change on agriculture, continuous monitoring and evaluations of agricultural variables is vital and remains critical for achieving long-term agricultural sustainability and food security \citep{becker-reshef_crop_2023, challinor_meta-analysis_2014,kang_climate_2009}. Accordingly, in view of the importance of agricultural variables across a wide range of applications, substantial advances in data analytics, remote sensing technologies, and Machine Learning (ML) algorithms have been leveraged for their assessment \citep{muruganantham_systematic_2022, pierre_pott_crop_2022}.

ML algorithms are widely employed for the prediction of different agricultural variables \citep{feng_crop_2019, rashid_comprehensive_2021, tariq_mapping_2023}. These algorithms are able to process vast amounts of heterogeneous data, including meteorological variables, soil characteristics, crop health indicators, and remote sensing data. In a broad sense, supervised ML models apply advanced computational strategies to analyze relationships between specific agricultural independent variables (e.g., soil moisture, climate conditions, crop growth stages) and specific agricultural dependent variable(s) such as crop yield and crop type. These models then apply the learned relationships to provide agricultural variables in spatial or temporal contexts where direct measurements are absent.

Supervised ML algorithms are inherently dependent on training data, as they learn patterns, relationships, and representations from examples, for developing a functional mapping between inputs and outputs \citep{jordan_machine_2015}. From a statistical learning perspective, having a larger training dataset helps the model approximate the underlying data distribution more accurately, which improves the model’s performance and its ability to generalize beyond the training set \citep{caro_generalization_2022, cui_effect_2018, raudys_small_1991}. However, it is important to note that data quantity alone does not guarantee optimal performance; the data fed to ML algorithms must also be of high quality and representative to ensure reliable outcomes \citep{elmes_accounting_2020, foody_toward_2004}. Thus, both quantity and quality of training data play a significant role in determining the success of a given algorithm in terms of predictive performance and robustness.

Despite technological advancements and the urgent need, the availability of agricultural reference data remains inadequate \citep{condran_machine_2022, khanal_remote_2020, weiss_remote_2020, weitkamp_evaluating_2023}. The deficiencies in data quality, quantity, and spatial-temporal coverage present significant obstacles that can hinder the development of state-of-the-art predictive ML models \citep{emmanuel_survey_2021, hoffmann_machine_2019, yan_incomplete_2016}. In many real-world agricultural applications, the available reference data also suffer from measurement inaccuracies, incomplete records, and limited representation \citep{calin_analysis_2023, phalke_mapping_2020, pham_crop_2022, pierre_pott_crop_2022}. Furthermore, inconsistencies in data collection methods and data characteristics hinder cross-regional and long-term comparative studies, mainly because agricultural variables differ substantially across regions and change over time in response to external factors such as changing climatic conditions and evolving agricultural systems.

To address the challenges associated with insufficient reference data across diverse research domains, various methodologies have been developed and applied to bridge these data gaps \citep{lu_machine_2025, pan_survey_2010}. These approaches can be broadly categorized into two main primary groups: algorithmic-centric and data-centric methods. Algorithmic-centric techniques focus on optimizing the framework to extract maximum value from incomplete reference data through techniques such as meta-learning and probabilistic modeling. On the other hand, data-centric methods focus on modifying, expanding, or refining reference data through techniques such as data augmentation, Synthetic Data Generation (SDG), and data integration from multiple sources.

Data-centric methodologies, especially SDG algorithms, offer advantages beyond enhancing ML models’ performance and generalization by directly addressing limitations in reference data. They are generally aiming at generating synthetic data that maintains key structural characteristics of the original data, to supplement incomplete data, thereby reducing reliance on costly, time-consuming, or sensitive data acquisition processes \citep{lu_machine_2025, paulin_review_2023}.

This process involves the application of mathematical and probabilistic models to replicate the underlying structure of the original data. SDG algorithms enable the creation of realistic yet synthetic data that can be used for various purposes including but not limited to model training-validation-testing, and scenario simulation \citep{de_melo_next-generation_2022, fonseca_tabular_2023}. Furthermore, these algorithms help protect sensitive information by providing a mechanism for generating privacy-preserving synthetic data that retain statistical properties of the original data \citep{jordon_pate-gan_2018, pezoulas_synthetic_2024}, facilitating data sharing and compliance with regulatory requirements.

In recent years, a variety of SDG algorithms has been introduced and applied across multiple domains \citep{gm_comprehensive_2020, hernandez_synthetic_2022, lu_machine_2025}. Many commonly used SDG approaches are designed around one of the two objectives of local interpolation in feature space or global distributional fidelity \citep{chawla_smote_2002, goodfellow_generative_2014,van_breugel_synthetic_2023}; however, neither objectives, by itself, guarantees that the generated synthetic data are useful for the downstream learning task.

A synthetic sample is more likely to be beneficial to a downstream learner when it is plausible under the real data distribution, informative under a calibrated predictive model, and diverse or non-redundant relative to the real data \citep{achterberg_fidelity-agnostic_2025, jamnik_tabebm_2024, xu_utility_2025}. On this basis, tabular SDG \citep{fonseca_tabular_2023} can be formulated as a supervised, task-conditioned problem rather than as an unconditional data-mimicking problem. The central assumption is that useful synthetic samples must satisfy two requirements simultaneously: they should be statistically plausible under the original data distribution, and they should preserve the feature–target relationship required by the downstream classifier or regressor.

To operationalize this assumption, and building on the theoretical grounding of teacher–student knowledge transfer \citep{hinton_distilling_2015} and in-context learning for tabular data \citep{hollmann_accurate_2025}, a Task-Conditioned Synthetic Data Generation (TCSDG) framework was proposed, composed of a Bayesian Network (BN) candidate generator, a foundation-model teacher acting as a task-aware critic, and a coverage-preserving selection mechanism.

The task-conditioning arises primarily from the teacher-guided filtering, the target-space budget allocation, and the coverage-preserving selection, rather than from the candidate generator alone. Concretely within the framework, BN algorithm \citep{ankan_pgmpy_2015, qian_synthcity_2023} generates a large candidate pool of synthetic samples, from which candidate samples are then evaluated by TabICL (Tabular Foundation Model for In-Context Learning) algorithm \citep{qu_tabicl_2025} as predictive teacher rather than accepted solely on the basis of generative likelihood or distributional similarity. The use of a tabular in-context learner as the teacher is motivated by recent evidence that tabular foundation models can perform strong supervised prediction by conditioning directly on labeled examples \citep{qu_tabicl_2025}. Furthermore, a mini-batch k-means diversification step and a convex budget allocation scheme balancing empirical and inverse-frequency priors ensure that minority classes and conditions receive proportionally enriched synthetic support without distorting the global marginal distribution.

The contributions of this work are threefold. First, this study introduces TCSDG, a task-conditioned pipeline for supervised tabular learning that integrates target-space budget allocation, teacher-guided consistency filtering, coverage-preserving selection, and quality-aware sample weighting. Second, TCSDG is benchmarked under identical experimental conditions against six established tabular SDG methods spanning the principal methodological families in the literature (probabilistic-graphical, adversarial, latent-variable, and diffusion-based), providing a controlled, utility-first comparison. Third, this comparison is instantiated on twelve supervised agricultural prediction tasks: six crop type classification datasets and six crop yield prediction datasets covering maize and wheat, providing a systematic multi-site, multi-task evaluation of tabular SDG utility in precision agriculture.

The multi-site, multi-task design is motivated by the well-documented sensitivity of data-driven agricultural models to environmental, soil, and climatic variability \citep{alami_machichi_crop_2023, ebrahimy_utilization_2023, van_klompenburg_crop_2020, wang_progress_2024}. By conducting experiments across multiple study sites and settings, this research seeks to ensure that performance evaluations are not site-specific but instead capture the variability inherent in different agricultural systems, thus improving the reliability and generalizability of the achieved conclusions regarding the impact of SDG algorithms.

\section{Materials and Methods}
\subsection{TCSDG (Task-Conditioned Synthetic Data Generation) Algorithm}

\begin{figure*}[!h]
    \centering \includegraphics[width=\linewidth,height=0.75\textheight,keepaspectratio]{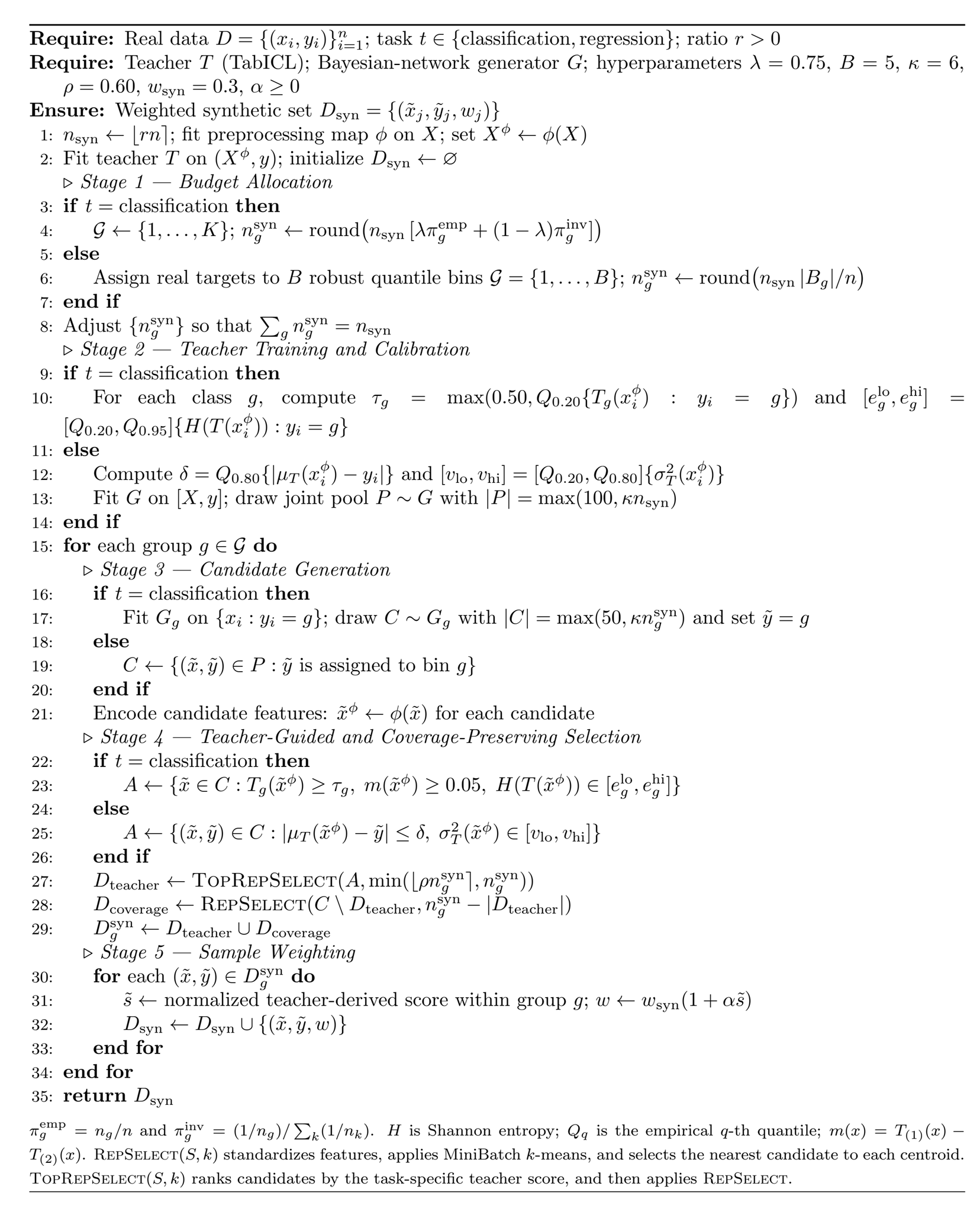}
    \caption{Pseudocode of the TCSDG algorithm for tabular synthetic data generation.}
    \label{fig:Pseudocode}
\end{figure*}

TCSDG is a task-conditioned synthetic data generation algorithm for supervised tabular learning. Figure~\ref{fig:Pseudocode} summarizes the complete TCSDG procedure as pseudocode. Given a real training set and a user-specified synthetic-to-real ratio r >0, TCSDG produces synthetic records that can be used either to replace the original training data and/or to augment it for a downstream supervised task. The algorithm proceeds in five stages: budget allocation, teacher training and calibration, candidate generation, teacher-guided and coverage-preserving selection, and sample weighting.

\subsubsection{Budget Allocation}

The purpose of the budget allocation step is to control how the total synthetic budget is distributed across the target space, such that the main structure of the real data distribution is preserved while controlled reinforcement of underrepresented regions is still possible.

For classification, let $n_c$ denote the number of real training samples in class $c$, where $c \in \{1, \dots, K\}$. The empirical class prior is defined as:
\begin{equation}
\pi_c^{\text{emp}} = \frac{n_c}{n}
\end{equation}

To increase the representation of rare classes, an inverse-frequency prior is also computed:
\begin{equation}
\pi_c^{\text{inv}} = \frac{1/n_c}{\sum_k 1/n_k}
\end{equation}

The final class allocation prior is defined as:
\begin{equation}
\pi_c = \lambda \pi_c^{\text{emp}} + (1 - \lambda)\, \pi_c^{\text{inv}}
\end{equation}

where $\lambda \in [0, 1]$ controls the trade-off between preserving the observed class distribution and increasing minority-class representation ($\lambda = 0.75$ by default).

For regression, the target variable is partitioned into $B$ equal-frequency bins ($B = 5$ by default) using quantile binning. Let $\mathcal{B}_b$ denote the set of training samples whose target values fall in bin $b$. The regression allocation follows the bin proportions:
\begin{equation}
\pi_b^{\text{reg}} = \frac{|\mathcal{B}_b|}{n}, \qquad n_b^{\text{syn}} = \operatorname{round}\!\left(n_{\text{syn}}\, \pi_b^{\text{reg}}\right)
\end{equation}

Equal-frequency binning was chosen because it spreads the synthetic budget uniformly across the response range without further parametric assumptions.

\subsubsection{Teacher Training and Calibration}
\label{sec:2.1.2}

TabICL \citep{qu_tabicl_2025} is used as the task-aware teacher.TabICL is an in-context, tabular foundation model that requires no per-task gradient-based tuning, which makes it well suited to settings where real training data are limited. In TCSDG, the teacher is fitted on the real training data and is then used only to evaluate candidate synthetic samples. The teacher therefore functions as a task-aware critic rather than a generator component.
For classification, the teacher returns class-probability estimates:

\begin{equation}
T(x) = \big(T_1(x), \dots, T_K(x)\big)
\end{equation}
where $T_c(x)$ denotes the predicted probability that $x$ belongs to class $c$. For a candidate generated with assigned class $c$, three quantities are computed: the assigned-class probability ($p_c(x) = T_c(x)$), the prediction margin ($m(x) = T_{(1)}(x) - T_{(2)}(x)$; where $T_{(1)}(x)$ and $T_{(2)}(x)$ are the largest and second-largest predicted class probabilities), and the predictive entropy ($H(T(x)) = -\sum_{k=1}^{K} T_k(x)\, \log T_k(x)$).

A classification candidate is considered teacher-consistent if it satisfies three criteria simultaneously: (1) its assigned-class probability satisfies $p_c(x) \geq \tau_c$, where the per-class threshold is $\tau_c = \max\!\big(0.50,\, Q_{0.20}\{p_c(x_i) : \text{real samples of class } c\}\big)$; (2) its margin satisfies $m(x) \geq 0.05$; (3) its entropy lies within the class-specific interval $[Q_{0.20}, Q_{0.95}]$ of the entropy values observed for real samples in class $c$. This prevents the selection of candidates that are either highly uncertain or unrealistically low-entropy relative to the real training data.

For regression, the teacher provides predictive mean $\mu_T(x)$ and predictive variance $\sigma_T^2(x)$. For a generated candidate $(\tilde{x}, \tilde{y})$, the teacher residual is defined as:
\begin{equation}
d(\tilde{x}, \tilde{y}) = |\mu_T(x) - \tilde{y}|
\end{equation}

All teacher thresholds are calibrated using predictions on the real training data. For classification, the calibration statistics are the per-class probability thresholds and entropy intervals. For regression, they are the residual threshold $\delta = Q_{0.80}\{|\mu_T(x_i) - y_i| : \text{real training sample}\}$ and the predictive-variance interval $[Q_{0.20}, Q_{0.80}]$ of $\sigma_T^2(x_i)$ over the real training samples.

\subsubsection{Candidate Generation}
\label{sec:2.1.3}
The candidate generation stage constructs an oversized pool of synthetic samples. The candidate pool is generated at least six times the required synthetic budget per class (classification) or in total (regression). The six times factor was chosen as a compromise between providing the teacher-consistency and coverage-selection stages with enough headroom to filter aggressively and keeping the computational cost manageable.

For classification, class-conditional BN generators are used. Let $D_c = \{x_i : (x_i, y_i) \in D,\ y_i = c\}$ denote the set of feature vectors belonging to class $c$. A separate generator $G_c$ is fitted to each class-specific feature subset $D_c$, thereby approximating $p(x \mid y = c)$. Synthetic classification candidates are generated as:
\begin{equation}
\tilde{x}_j^{(c)} \sim G_c, \qquad \tilde{y}_j^{c} = c
\end{equation}

For regression, a single BN generator ($G$) is fitted to the combined feature-target data $[X, y]$, and candidates are sampled as:
\begin{equation}
\tilde{x}_j, \tilde{y}_j \sim G
\end{equation}
This joint formulation avoids fragmenting the training data into small target-specific subsets and preserves local feature-response structure before teacher-based filtering is applied.
\subsubsection{Teacher-Guided and Coverage-Preserving Selection}
\label{sec:2.1.4}
The final synthetic dataset is selected from the candidate pool as the union of two disjoint subsets:
\begin{equation}
D_{\text{syn}} = D_{\text{teacher}} \cup D_{\text{coverage}}
\end{equation}

The teacher-consistent subset $D_{\text{teacher}}$ contains candidates satisfying the task-specific teacher criteria described in Section~\ref{sec:2.1.2}. This subset promotes predictive relevance by excluding candidates with implausible labels, weak class support, excessive uncertainty, or extreme feature--target disagreement. The coverage subset $D_{\text{coverage}}$ is selected from the remaining candidate pool with minimal teacher intervention, preserving distributional support and reducing dependence on the teacher model.

Within each class for classification, or within each target bin for regression, a fraction of the local budget is allocated to teacher-consistent samples; the remaining budget is filled by coverage samples.

Teacher-consistent candidates are first ranked by a task-specific teacher score, and only the highest-ranked subset is passed to representative selection. Coverage samples are selected from the remaining candidate pool without enforcing the teacher-consistency constraints. In both cases, representative selection standardizes the candidate features, applies MiniBatch $k$-means with the required number of clusters, and returns the candidate closest to each centroid. This procedure promotes diversity while keeping the computational cost manageable.

For classification, the selected synthetic subset for class $c$ satisfies:
\begin{equation}
D_{\text{syn}}^{(c)} = D_{\text{teacher}}^{(c)} \cup D_{\text{coverage}}^{(c)}, \qquad \big|D_{\text{syn}}^{(c)}\big| = n_c^{\text{syn}}
\end{equation}

For regression, equal-frequency target bins computed from the real training targets are used to allocate the synthetic budget. Candidate samples are then organized by rank-based target-bin identifiers within the generated candidate pools, and the same teacher-consistency and representative-selection procedure is applied within each target bin $b$ so that $\big|D_{\text{syn}}^{(b)}\big| = n_b^{\text{syn}}$.

\subsubsection{Sample Weighting}
\label{sec:2.1.5}
TCSDG provides a sample-weighting mechanism for synthetic samples, which can be used to assign higher importance to the original data and preserve the relative contribution of real data in the downstream tasks. For this purpose, let $w_{\text{syn}} \leq 1$ denote the base synthetic weight that after some preliminary analyses set at $0.3$. For each selected synthetic sample, a normalized teacher-derived score $\tilde{s}_j \in [0, 1]$ is computed within the corresponding class or target bin. The final synthetic weight is defined as:
\begin{equation}
w_j = w_{\text{syn}}\left(1 + \alpha \tilde{s}_j\right)
\end{equation}

where $\alpha \geq 0$ controls the strength of the teacher-based weight adjustment. For classification, $\tilde{s}_j$ is derived from the assigned-class probability, the margin, and the entropy deviation. For regression, teacher-consistent candidates are ranked using residual agreement and the predictive-variance deviation, then $\tilde{s}_j$ is computed from a normalized residual agreement score, implemented as the negative teacher residual within each target bin.

The weighted empirical objective for a downstream model $f$ is:
\begin{equation}
\min_{f} \sum_{(x_i, y_i) \in D} l\big(f(x_i), y_i\big) + \sum_{(\tilde{x}_i, \tilde{y}_i) \in D_{\text{syn}}} w_i\, l\big(f(\tilde{x}_i), \tilde{y}_i\big)
\end{equation}

where $l(\cdot, \cdot)$ denotes the task-specific loss function.

\subsection{Overall Experimental Procedures}
To evaluate the proposed TCSDG algorithm and to position it relative to existing tabular SDG methods, a controlled benchmarking procedure was designed with two complementary objectives: first, to quantify the practical utility of TCSDG-generated synthetic data when used to augment training sets for downstream tasks; and second, to compare its utility, under identical conditions, against a set of established SDG algorithms.

The evaluation was organized around five research questions: (i) Does augmenting real training data with TCSDG-generated samples improve downstream predictive performance over training on real data alone? (ii) Can synthetic data generated by TCSDG, used in place of real training data, match or exceed the performance obtained with real training data? (iii) Does TCSDG outperform established tabular SDG baselines under the same experimental protocol? (iv) Is the utility of TCSDG stable across heterogeneous country-level agricultural datasets, or is it driven by a few favourable cases? (v) How do the synthetic-to-real multiplication ratio, the size of the real training set, and the choice of downstream ML learner shape the utility of the generated data?

The benchmark comprised twelve independent supervised-learning tasks; including six regression tasks targeting crop yield prediction (maize in Argentina, Germany, and Zambia; wheat in India, France, and Italy), and six classification tasks targeting crop type classification in Greece, Italy, Austria, Germany, Romania, and Bulgaria. For every task, the original reference dataset was treated as the ground-truth source and partitioned using a standardized random-splitting strategy. Each dataset was divided into a training partition (70\%) and a validation partition (30\%); for classification tasks the split was stratified by class label to preserve class proportions. The validation partition was withheld entirely from both SDG and ML model training, and was used exclusively for the final performance assessment. To account for sampling variability and to mitigate the effect of any single random partition, the 70/30 split, together with the full downstream pipeline that depends on it, was repeated ten times per task using ten fixed random seeds. All reported results are therefore expressed as the mean and standard deviation over these ten repetitions.

To examine performance under different levels of data availability, two training regimes were derived from training partition: a full regime (TD70), comprising the complete 70\% training partition, and a reduced regime (TD30), comprising a random subsample equal to 30\% of the full dataset. Within each regime, synthetic samples were generated at four multiplication ratios, 1× (singlet), 2× (doubled), 4× (quadrupled), and 8× (octupled), defined relative to the number of real training samples.

Synthetic data were produced by TCSDG and, for benchmarking, by six established tabular SDG algorithms drawn from the Synthcity python library \citep{qian_synthcity_2023}. The benchmark SDG algorithms were BN \citep{ankan_pgmpy_2015}, Conditional Tabular Generative Adversarial Networks (CTGAN) \citep{xu_modeling_2019}, Anonymization through Data Synthesis using Generative Adversarial Networks (ADSGAN) \citep{yoon_anonymization_2020}, Tabular Variational Autoencoders (TVAE) \citep{xu_modeling_2019}, Robust TVAE (RTVAE) \citep{akrami_robust_2022}, and tabular Denoising Diffusion Probabilistic Model (DDPM) \citep{kotelnikov_tabddpm_2023}.

These six SDG models were selected because they provide sufficiently representative coverage of the major methodological families in tabular SDG. Specifically, the BN represents probabilistic graphical and dependency-based synthesis; CTGAN and ADSGAN represent adversarial generation, with ADSGAN additionally incorporating a privacy-aware synthesis objective; TVAE and RTVAE represent latent-variable VAE-based synthesis, including a robustness-oriented variant; and DDPM represents the class of diffusion-based generators for heterogeneous tabular data.

Three ML algorithms were employed for both use cases: Random Forest (RF) \citep{breiman_random_2001}, Multilayer Perceptron (MLP) \citep{hornik_multilayer_1989}, and Support Vector Machine (SVM) \citep{hearst_support_1998}. RF, SVM, and MLP were selected due to their theoretical significance and their ability to encapsulate a broad spectrum of ML algorithms. RF represents an ensemble-based approach that leverage decision trees, SVM provides a strong basis for margin-based optimization and kernel methods, and MLP captures the essence of neural network architectures.

To isolate the contribution of synthetic data, each downstream ML algorithm was trained under three configurations: (1) baseline, using the real training subsets (TD30 and TD70) without augmentation, which served as the baseline; (2) synthetic, using exclusively the generated synthetic samples at each regime and ratio; and (3) merged, using the combination of real and synthetic samples.

This design yielded, per task and per SDG method, six bassline configurations (3 ML learners × 2 training data regimes), 24 synthetic configurations (3 ML learners × 2 training data regimes × 4 multiplication ratios), and 24 merged configurations (3 ML learners × 2 training data regimes × 4 multiplication ratios). In the merged configuration, real samples received unit weight, while, for TCSDG, synthetic samples additionally carried quality-aware instance weights (Section~\ref{sec:2.1.5}) derived from the teacher model. The weights are utilized only by downstream ML learners that support weighted training (RF and SVM).

The complete implementation and the benchmarking code are publicly available at \href{https://github.com/HamidEbrahimy/TCSDG}{https://github.com/HamidEbrahimy/TCSDG}.

\subsection{Case studies}
\subsubsection{Crop Type Classification}
European Land Use/Cover Area frame Survey (LUCAS) dataset 2022 \citep{dandrimont_advances_2024} that provides a unique source of publicly available land cover survey data, was used for the crop type classification task. LUCAS collects information on land cover and land use, agro-environmental, and soil data every three years. The collected in-situ information is published and is accompanied by detailed metadata, such as the acquisition date and GPS location of the observer \citep{dandrimont_advances_2024}. Although the main purpose of LUCAS is statistical estimation and it was not designed for mapping, the potential and thoroughness of it leads to its applications in several classification tasks \citep{pflugmacher_mapping_2019, weigand_spatial_2020}.

The LUCAS 2022 dataset was specifically used, and the crop classes were selected from the provided hierarchical classification scheme. The data were then stratified by country, as agricultural landscapes generally vary across regions in response to both biophysical conditions and human management decisions \citep{andersen_farming_2017}. Evaluating the crop type classification procedure separately for each country enabled us to account for this heterogeneity and reduced the risk that classification accuracy would be biased by dominant data sources or region-specific farming practices. In addition, crop classes represented by fewer than 50 samples were removed from each country-specific dataset.

\begin{figure*}[!h]
    \centering \includegraphics[width=\linewidth,height=0.8\textheight,keepaspectratio]{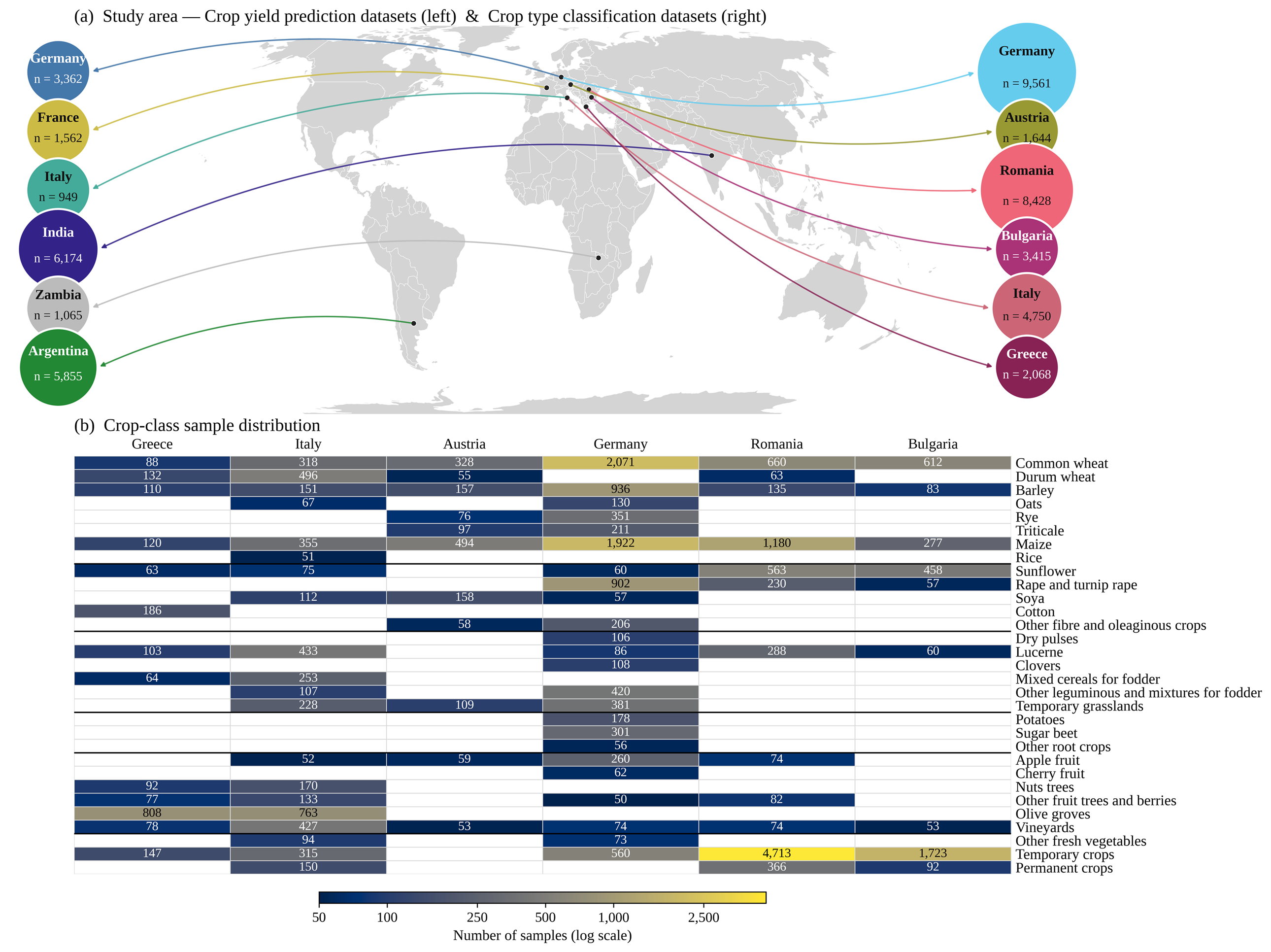}
    \caption{Benchmark dataset overview. (a) Study areas for the two tasks: crop yield prediction (Germany, France, Italy, India, Zambia, and Argentina) and crop type classification (Germany, Austria, Romania, Bulgaria, Italy, and Greece). Circle size is proportional to the number of samples (n) in each country-level dataset. (b) Per-class sample counts for the crop type classification datasets that displayed on a logarithmic color scale. White cells indicate the absence of a crop class in a given country.}
    \label{fig:Study area}
\end{figure*}

Among the available countries, Greece, Italy, Austria, Germany, Romania, and Bulgaria were selected. The crop classes, the number of samples per class, and the total sample size available for each selected country are presented in Figure~\ref{fig:Study area}. The class distribution clearly indicates that crop type classification is a highly imbalanced classification problem.

To employ the LUCAS data for the crop type classification task, multi-temporal Sentinel-2 data were used as predictor variables. Specifically, for each LUCAS sample, all Sentinel-2 images from 2022 with cloud coverage below 20\% were acquired. From the Sentinel-2 spectral bands, ten bands were extracted: Band 2 (Blue), Band 3 (Green), Band 4 (Red), Band 5 (Red Edge 1), Band 6 (Red Edge 2), Band 7 (Red Edge 3), Band 8 (NIR), Band 8A (Red Edge 4), Band 11 (SWIR 1), and Band 12 (SWIR 2). For each band, three temporal metrics (minimum, median, and maximum) were then calculated to generate a Sentinel-2 composite. Accordingly, 30 temporal metrics of Sentinel-2 pixels corresponding to the locations of the LUCAS samples were extracted; these comprised the per-pixel minimum, median, and maximum values for each of the ten spectral bands.
\subsubsection{Crop Yield Prediction}

Crop yield prediction experiments were carried out using version 1.4 of CY-Bench dataset \citep{paudel_cy-bench_2024}, which focuses on subnational, end-of-season yield data for maize and wheat. It includes information from both major crop-producing nations and underrepresented regions around the world. The dataset integrates subnational crop yield data with relevant predictors including weather during the growing season, remote sensing data, evapotranspiration, soil moisture, and static soil characteristics. CY-Bench was specifically curated to support model comparison across varied global agricultural systems in a way that closely mirrors real-world scenarios \citep{paudel_cy-bench_2025}.

Among the country-specific data available in the CY-Bench dataset, data from Argentina, Germany, and Zambia were used for maize yield prediction, while for wheat yield prediction, data from India, France, and Italy were used (Figure~\ref{fig:Study area}). These country subsets were selected to span a broad range of sample sizes (from 949 to 6,174 records) and to represent distinct agricultural and climatic systems.

For each country, the CY-Bench dataset consists of static and multi-temporal predictor variables, of which the available water capacity, bulk density, and drainage class are the static variables, while the time-series variables include the fraction of absorbed photosynthetically active radiation, average temperature, maximum temperature, minimum temperature, precipitation, radiation, potential evapotranspiration, climatic water balance, normalized difference vegetation index, surface soil moisture, rootzone soil moisture.

Considering the large number of predictor variables in the CY-Bench dataset, data preprocessing was performed. Specifically, time-series variables outside the period between the start and end of the growing season were excluded. Then, for each time-series variable, the minimum, median, and maximum values were calculated across the growing season. Moreover, the variable named “minimum precipitation” was removed because it was effectively zero in most cases. The final set of predictor variables for each country consisted of 35 predictor variables.

\subsection{Performance Evaluation}
The SDG algorithms were evaluated based on their ability in providing synthetic data that is capable to improve the predictive performance of the ML algorithms in crop yield prediction and crop type classification. All performance assessment procedures were deployed on the held-out validation partition which was unseen by both SDG and ML training procedures, and augmentation effectiveness was quantified by comparing ML model performance with and without synthetic data.

In the case of the crop type classification task, considering the imbalanced nature of the crop classes in the utilized reference datasets, the Balanced Accuracy (BAC) score \citep{velez_balanced_2007} was considered for performance evaluation. This metric is specifically designed to compensate for differences in sample sizes among classes.
\begin{equation}
\text{BAC} = \frac{1}{K} \sum_{k=1}^{K} \frac{\text{TP}_k}{\text{TP}_k + \text{FN}_k}
\end{equation}

where $K$ is the number of classes, $\text{TP}_k$ is the number of correctly classified samples of class $k$, and $\text{FN}_k$ is the number of samples belonging to class $k$ but classified as another class.

On the other hand, for the crop yield prediction task, Root Mean Squared Error (RMSE) was utilized as the performance evaluation metric.
\begin{equation}
\text{RMSE} = \sqrt{\frac{1}{n} \sum_{i=1}^{n} (\hat{y}_i - y_i)^2}
\end{equation}

where $\hat{y}_i$ and $y_i$ denote the predicted and the observed value for observation $i$, respectively, and $n$ is the total number of observations.

Moreover, to determine whether the differences between each SDG regime and the baseline were statistically meaningful, paired non-parametric significance testing was conducted. For each experimental setting, the regime under consideration was paired with the baseline on a per-repetition basis, and differences were calculated using the task-appropriate metric. Specifically, the two-sided Wilcoxon signed-rank test was applied to evaluate the null hypothesis that the median paired difference was zero without relying on parametric assumptions. To control the family-wise error rate across the multiple methods evaluated within each task, raw p-values were adjusted using the Holm step-down procedure. A comparison was considered significant only if its Holm-adjusted p-value was below 0.05 and the mean difference indicated improved ML model performance for the task under consideration.
\section{Results and Discussion}
\subsection{Overall Utility of TCSDG}
\label{sec:3.1}
TCSDG was the only SDG method that produced statistically reliable improvements over the no-augmentation baseline for all the twelve crop type classification and crop yield prediction tasks (Table ~\ref{tab:winrate}). In the merged regime, TCSDG outperformed the baseline in 89\% of the classification comparisons and 74\% of the regression comparisons. Among the six benchmark generators, only BN achieved a significant improvement, and only for crop yield prediction in the merged regime. The relative competitiveness of BN is also instructive, as TCSDG itself uses BN-based candidate generation, but the plain BN benchmark was far weaker than TCSDG, especially for classification. This indicates that TCSDG’s advantage cannot be attributed simply to the BN generator. Rather, the main contribution lies in the task-conditioned framework surrounding it. On the other hand, CTGAN, ADSGAN, TVAE, RTVAE, and DDPM did not provide significant positive utility in either task.

The synthetic regime provides a stricter test of whether the generated data independently preserved task-relevant structure. TCSDG again performed best compared to the other SDG methods. For crop type classification, TCSDG synthetic-only regime exceeded the baseline in 91\% of paired comparisons and achieved a significant mean BAC gain of 0.05. This result indicates that TCSDG did not merely add volume to the training set; it generated samples that retained sufficient discriminative information to train effective classifiers without access to the original records. In crop yield prediction, TCSDG synthetic-only regime were less effective, winning 44\% of comparisons and showing a small and non-significant mean RMSE improvement. However, this was still the strongest synthetic-only result among all evaluated methods in crop yield prediction task, whereas all other non-TCSDG generators failed almost completely in synthetic-only regime.

\begin{table*}[!h]
\small
\centering
\caption{Aggregated utility of TCSDG and six benchmark SDG methods, pooled across datasets, training fractions, multiplication ratios, downstream learners, and repetitions. Values shown are for crop type classification with crop yield prediction in parentheses. The Win Rate column shows the proportion of paired runs in which the regime outperformed the baseline. The Win Margin column shows the mean paired difference relative to the baseline (BAC for classification, with sign-flipped RMSE for regression). The bold values show the most accurate method, and the entries marked with * indicate that the Holm-corrected, two-sided Wilcoxon signed-rank test p-value is below 0.05 and the mean difference is in the direction of improvement.}
\label{tab:winrate}
\begin{tabular}{lcccc}
\toprule
\multirow{2}{*}{Methods} & \multicolumn{2}{c}{Win Rate (\%)} & \multicolumn{2}{c}{Win Margin} \\
\cmidrule(lr){2-3} \cmidrule(lr){4-5}
 & Merged & Synthetic & Merged & Synthetic \\
\midrule
TCSDG & \textbf{89 (74)} & \textbf{91 (44)} & $\mathbf{+0.04^*}$ ($\mathbf{+0.02^*}$) & $\mathbf{+0.05^*}$ ($\mathbf{+0.01}$) \\
CTGAN  & 12 (14) & 2 (0)    & $-0.02$ ($-0.04$)     & $-0.08$ ($-0.39$) \\
ADSGAN & 12 (15) & 2 (0)    & $-0.02$ ($-0.04$)     & $-0.08$ ($-0.37$) \\
DDPM   & 6 (11)  & 3 (0)    & $-0.04$ ($-0.17$)     & $-0.08$ ($-2.50$) \\
BN     & 26 (68) & 10 (35)  & $-0.02$ ($+0.01^*$)   & $-0.05$ ($-0.02$) \\
TVAE   & 17 (21) & 2 (0)    & $-0.02$ ($-0.02$)     & $-0.06$ ($-0.32$) \\
RTVAE  & 24 (21) & 0.3 (0)  & $-0.01$ ($-0.03$)     & $-0.08$ ($-0.75$) \\
\bottomrule
\end{tabular}
\end{table*}

These findings support the central premise of the proposed algorithm: synthetic data are useful only when they preserve the conditional structure required by the downstream task. The six benchmark generators are primarily designed to approximate the joint distribution of a tabular data, whereas TCSDG adds task-conditioned budget allocation, teacher-guided filtering, coverage-preserving selection, and quality-aware weighting. This distinction is critical for agricultural prediction, where utility depends not simply on realistic records but on preserving the relationships between predictors and crop labels or yield values. This interpretation is consistent with the broader tabular SDG literature, which emphasizes that mixed feature types, heterogeneous value ranges, and inter-feature correlations make tabular generation difficult \citep{anshelevich_synthetic_2025, fonseca_tabular_2023}. It also agrees with evidence that conventional tabular SDG methods do not consistently improve downstream ML efficacy in practical augmentation settings \citep{manousakas_usefulness_2023}.

The results therefore argue for a utility-first evaluation of agricultural SDG. In this study, a generator was judged by whether it improved prediction on held-out real validation data, not by intrinsic fidelity alone. This is consistent with the ML-efficiency paradigm used in tabular SDG, where synthetic data are evaluated by training discriminative models on generated records and testing them on real records \citep{borisov_language_2023}. It is also consistent with agricultural synthetic-data evidence showing that realism and utility are not equivalent; task-discriminative information can matter more than photorealism or superficial fidelity \citep{klein_synthetic_2024}. This task-utility-first framing is also adopted in recent work on fidelity-agnostic SDG, which explicitly argues that high statistical fidelity is neither necessary nor sufficient for downstream task utility and may in some cases harm privacy without aiding prediction \citep{achterberg_fidelity-agnostic_2025}.

\subsection{Effect of Synthetic Multiplication Ratio}
\label{sec:3.2}
\begin{figure*}[!h]
    \centering \includegraphics[width=\linewidth,height=0.8\textheight,keepaspectratio]{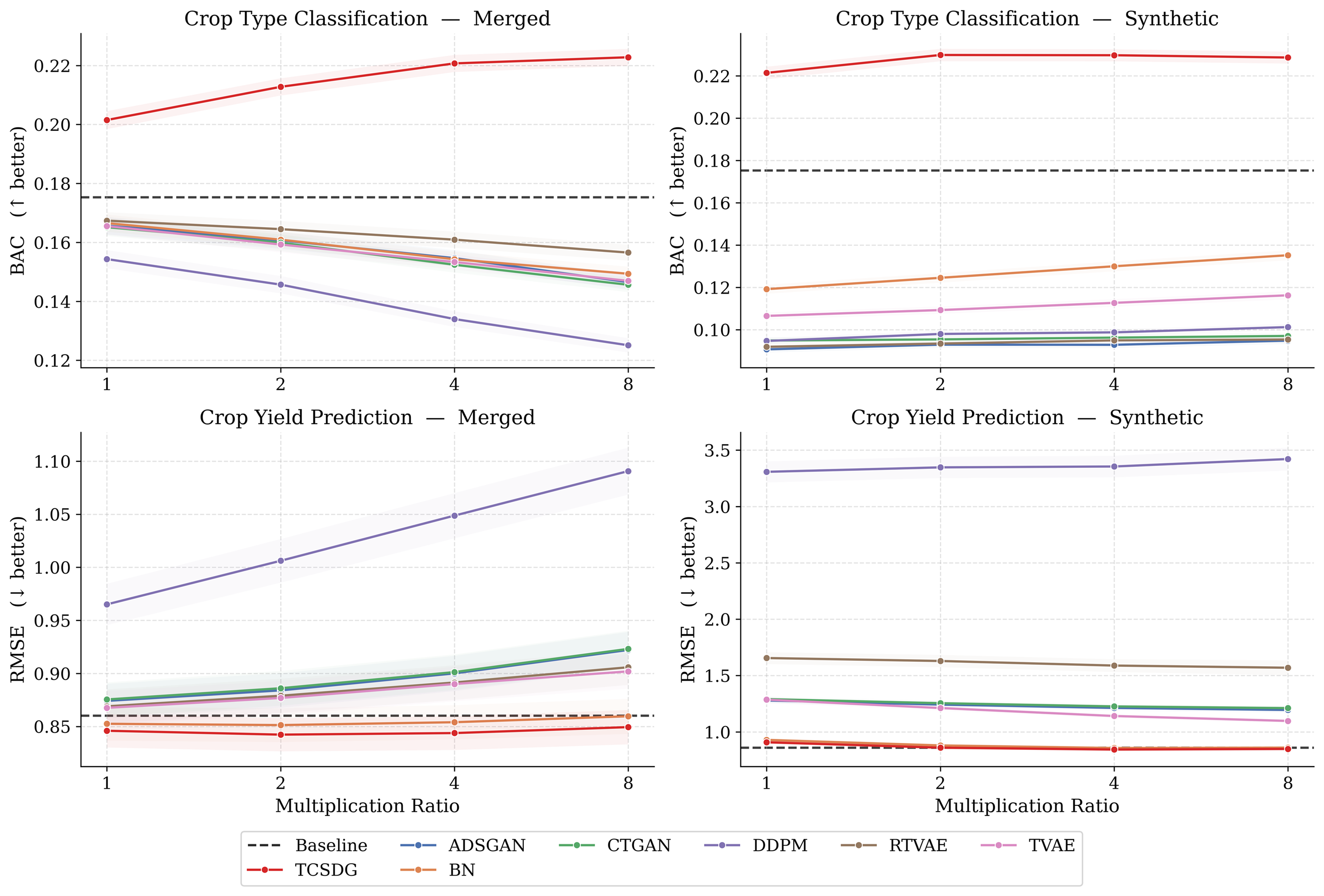}
    \caption{Joint effect of the synthetic multiplication ratio and downstream regime on ML performance, separated by task (classification: BAC, higher is better; regression: RMSE, lower is better) and regime (merged: synthetic + real; synthetic: synthetic only). The dashed black line is the no-augmentation baseline. Shaded bands are ±1 SD across the ten repetitions, pooled across the six datasets per task.}
    \label{fig:RatioImapct}
\end{figure*}

Figure~\ref{fig:RatioImapct} summarizes the impact of the multiplication ratio, defined as the relative size of the generated synthetic data, on downstream ML performance across tasks and training regimes. In the merged regime, the multiplication ratio effectively determines the relative contribution of synthetic samples to the combined real–synthetic training set. For all benchmark SDG methods, increasing this ratio was consistently detrimental. In crop type classification, BAC declined as the synthetic fraction increased, whereas in crop yield prediction, RMSE increased with each successive ratio. Thus, augmentation methods that were already below the real-data baseline at a ratio of 1 degraded further as the synthetic component became more dominant. This behavior indicates that, for these methods, additional synthetic samples diluted rather than enriched the real training distribution.

TCSDG was the only exception to this trend. In crop type classification, TCSDG remained above the baseline at all multiplication ratios in the merged regime. Its performance increased from ratios 1 to 4 and then saturated between ratios 4 and 8, suggesting that most of the classification benefit was achieved with a moderate synthetic budget. For crop yield prediction, TCSDG maintained stable RMSE values that remained below the baseline even as the multiplication ratio increased, indicating that additional synthetic samples were beneficial or, at worst, neutral.

The synthetic-only regime showed a different pattern. When real samples were removed from the training set, additional synthetic data was weakly beneficial or neutral for several benchmark SDG methods. However, their performance still plateaued well below the real-data baseline, indicating that larger synthetic datasets alone were insufficient to recover the task-relevant structure present in the original data. DDPM remained particularly the most ineffective, with consistently low classification performance and high yield-prediction error. The GAN- and VAE-based models, including ADS-GAN, CTGAN, TVAE, and RTVAE, as well as the BN, also clustered below the baseline, especially for crop type classification. TCSDG showed a similar saturation pattern, but at a substantially higher performance level. In crop type classification, peak BAC was reached by the doubled ratio, whereas in crop yield prediction, RMSE converged toward the real-data baseline with same ratio.

These findings indicate that the multiplication ratio should not be treated as a neutral scaling parameter. When synthetic samples are weakly aligned with the target task, increasing their number can amplify error rather than improve generalization. The deterioration observed for CTGAN, ADSGAN, TVAE, RTVAE, BN, and DDPM at larger ratios suggests that their samples increasingly dominated the training mixture without providing equivalent task-relevant information. This downward trend with increased synthetic fraction has also been observed in recent tabular benchmarks where SDG models supply samples that are statistically realistic but task-misaligned, producing net negative augmentation utility \citep{jamnik_tabebm_2024, manousakas_usefulness_2023}.

The robustness of TCSDG can be attributed to its constrained sample-generation strategy. Its final synthetic set was not drawn directly from an unconstrained generator. Instead, candidate samples were first generated in excess, then filtered according to teacher consistency, and finally selected to preserve coverage across crop classes or target bins. This procedure prevented performance collapse at larger ratios and produced controlled saturation rather than unlimited improvement. From a practical perspective, this result suggests that moderate augmentation ratios may be sufficient to capture most of the benefit of TCSDG, whereas excessive synthetic expansion is unnecessary.

\subsection{Downstream Learner Response}
\label{sec:3.3}
To test whether the conclusions of the aggregate analysis hold across individual downstream ML learners, Table ~\ref{tab:MLimpact} reports a per-learner breakdown of generator utility. These results confirm that the benefit of TCSDG was, in general, consistent in direction but that its magnitude depended on the learner it was paired with.

\begin{table*}[!h]
\small
\centering
\caption{Per-ML-learner breakdown of generator utility, pooled across datasets, training fractions, multiplication ratios, and repetitions. Win Rate (\%) is the proportion of paired runs in which the generator-and-learner combination outperformed the no-augmentation baseline. Performance Gain is the mean signed difference from that baseline, expressed in the natural metric units, where positive values denote improvement (BAC, for classification and sign-flipped RMSE for regression). The bold values show the most accurate method in each configuration, and the entries marked * are those that are statistically significant.}
\label{tab:MLimpact}
\footnotesize
\setlength{\tabcolsep}{5pt}
\begin{tabular}{l l l ccc ccc}
\toprule
& \multirow{2}{*}{Method} & \multirow{2}{*}{Mode}
& \multicolumn{3}{c}{Win Rate (\%)} & \multicolumn{3}{c}{Performance Gain} \\
\cmidrule(lr){4-6} \cmidrule(lr){7-9}
& & & MLP & RF & SVM & MLP & RF & SVM \\
\midrule
\multirow{14}{1.8cm}{Crop Type Classification}
 & \multirow{2}{*}{TCSDG}  & Merged    & \textbf{68} & \textbf{99} & \textbf{100} & \textbf{+0.011*} & \textbf{+0.037*} & \textbf{+0.069*} \\
 &                         & Synthetic & \textbf{75} & \textbf{99} & \textbf{100} & \textbf{+0.018*} & \textbf{+0.048*} & \textbf{+0.090*} \\
\cmidrule(lr){2-9}
 & \multirow{2}{*}{ADSGAN} & Merged    &  3 & 19 & 14 & $-$0.036 & $-$0.010 & $-$0.009 \\
 &                         & Synthetic &  0 &  1 &  4 & $-$0.131 & $-$0.068 & $-$0.049 \\
\cmidrule(lr){2-9}
 & \multirow{2}{*}{BN}     & Merged    &  8 & 22 & 47 & $-$0.044 & $-$0.011 & +0.003* \\
 &                         & Synthetic &  3 &  6 & 22 & $-$0.082 & $-$0.040 & $-$0.023 \\
\cmidrule(lr){2-9}
 & \multirow{2}{*}{CTGAN}  & Merged    &  7 & 16 & 13 & $-$0.038 & $-$0.011 & $-$0.009 \\
 &                         & Synthetic &  0 &  0 &  4 & $-$0.128 & $-$0.065 & $-$0.046 \\
\cmidrule(lr){2-9}
 & \multirow{2}{*}{DDPM}   & Merged    &  7 & 11 &  1 & $-$0.052 & $-$0.021 & $-$0.034 \\
 &                         & Synthetic &  0 &  1 &  9 & $-$0.119 & $-$0.063 & $-$0.049 \\
\cmidrule(lr){2-9}
 & \multirow{2}{*}{RTVAE}  & Merged    &  7 & 31 & 36 & $-$0.029 & $-$0.004 & $-$0.006 \\
 &                         & Synthetic &  0 &  0 &  1 & $-$0.129 & $-$0.068 & $-$0.047 \\
\cmidrule(lr){2-9}
 & \multirow{2}{*}{TVAE}   & Merged    &  5 & 20 & 27 & $-$0.040 & $-$0.011 & $-$0.006 \\
 &                         & Synthetic &  0 &  0 &  7 & $-$0.106 & $-$0.053 & $-$0.033 \\
\midrule
\multirow{14}{1.8cm}{Crop Yield Prediction}
 & \multirow{2}{*}{TCSDG}  & Merged    & \textbf{46} & \textbf{80} & \textbf{97} & \textbf{+0.008}  & \textbf{+0.015*} & \textbf{+0.028*} \\
 &                         & Synthetic & \textbf{36} & \textbf{34} & \textbf{60} & \textbf{$-$0.006} & \textbf{$-$0.017} & \textbf{+0.011*} \\
\cmidrule(lr){2-9}
 & \multirow{2}{*}{ADSGAN} & Merged    & 32 &  3 &  8 & $-$0.032 & $-$0.057 & $-$0.023 \\
 &                         & Synthetic &  0 &  0 &  0 & $-$0.427 & $-$0.385 & $-$0.302 \\
\cmidrule(lr){2-9}
 & \multirow{2}{*}{BN}     & Merged    & 36 & 75 & 93 & $-$0.008 & +0.014* & +0.021* \\
 &                         & Synthetic & 22 & 28 & 56 & $-$0.037 & $-$0.030 & +0.007  \\
\cmidrule(lr){2-9}
 & \multirow{2}{*}{CTGAN}  & Merged    & 30 &  2 &  7 & $-$0.037 & $-$0.060 & $-$0.024 \\
 &                         & Synthetic &  0 &  0 &  0 & $-$0.441 & $-$0.403 & $-$0.317 \\
\cmidrule(lr){2-9}
 & \multirow{2}{*}{DDPM}   & Merged    &  9 & 21 &  0 & $-$0.205 & $-$0.021 & $-$0.256 \\
 &                         & Synthetic &  0 &  0 &  0 & $-$2.994 & $-$1.690 & $-$2.268 \\
\cmidrule(lr){2-9}
 & \multirow{2}{*}{RTVAE}  & Merged    & 22 & 10 & 31 & $-$0.052 & $-$0.025 & $-$0.015 \\
 &                         & Synthetic &  0 &  0 &  0 & $-$1.063 & $-$0.502 & $-$0.397 \\
\cmidrule(lr){2-9}
 & \multirow{2}{*}{TVAE}   & Merged    & 24 &  4 & 34 & $-$0.036 & $-$0.039 & $-$0.009 \\
 &                         & Synthetic &  0 &  0 &  0 & $-$0.449 & $-$0.313 & $-$0.229 \\
\bottomrule
\end{tabular}
\end{table*}

For crop type classification, TCSDG was the only generator to improve all three ML algorithms in both the merged and synthetic regimes, with near-perfect win rates for RF and SVM (99–100\%) and a lower but still positive rate for MLP (68–75\%). The performance gains were largest for SVM (+0.069 merged, +0.090 synthetic), followed by RF and MLP. The same ordering held for crop yield prediction: TCSDG significantly improved SVM in both regimes and RF in the merged regime, whereas its effect on MLP was weak or slightly negative. Among the competing generators, only BN produced any positive performance gain, strongest with RF and SVM in the merged regime (+0.014, +0.021); every other SDG method negatively impacted RMSE relative to the baseline.

This learner-level view also sharpens the contrast between the two augmentation regimes. For the benchmark SDG methods, utility largely collapsed once the real training data were removed: synthetic-only win rates fell to or near 0\% in most regression settings, and the associated errors grew by an order of magnitude or more for the diffusion- and VAE-based methods (for example, DDPM reached mean changes of -2.994, -1.690, and -2.268 for MLP, RF, and SVM, respectively).

These learner-dependent patterns are consistent with both the algorithmic design of TCSDG and the inductive biases of the learners. In the merged regime, TCSDG synthetic samples carried teacher-derived, quality-aware weights, but these weights could only be exploited by learners that support weighted training, namely RF and SVM. Both could therefore up-weight high-confidence synthetic records while discounting lower-confidence ones, whereas MLP received no such signal, a plausible contributor to its weaker response. The effect was strongest for SVM because its decision boundary is determined by a subset of margin-defining support vectors \citep{mountrakis_support_2011}: well-placed synthetic points can directly improve the margin geometry. The teacher-consistency filter, by preferentially retaining the high-confidence samples that sharpen the margin and discarding those that would blur it, is the most likely reason SVM benefited most.

The comparative robustness of RF is partly consistent with the expectation that tree ensembles tolerate noisy inputs \citep{belgiu_random_2016}. Under the weakest classification generator, DDPM in the merged regime, RF showed the smallest BAC degradation (–0.021) relative to SVM (–0.034) and MLP (–0.052). The weaker MLP response should not, however, be read as evidence that neural models cannot benefit from synthetic data; rather, on these tabular agricultural tasks the MLP was simply more sensitive to imperfect synthetic calibration and could not access TCSDG's sample-weighting mechanism. Taken together, the per-learner results reinforce the broader observation that synthetic-data utility is determined jointly by generator quality, the downstream learner, and the augmentation size, rather than by any single factor \citep{ebrahimy_utilization_2023}.

\subsection{Effect of Training Data Size}
\label{sec:3.4}
\begin{figure*}[!h]
    \centering \includegraphics[width=\linewidth,height=0.8\textheight,keepaspectratio]{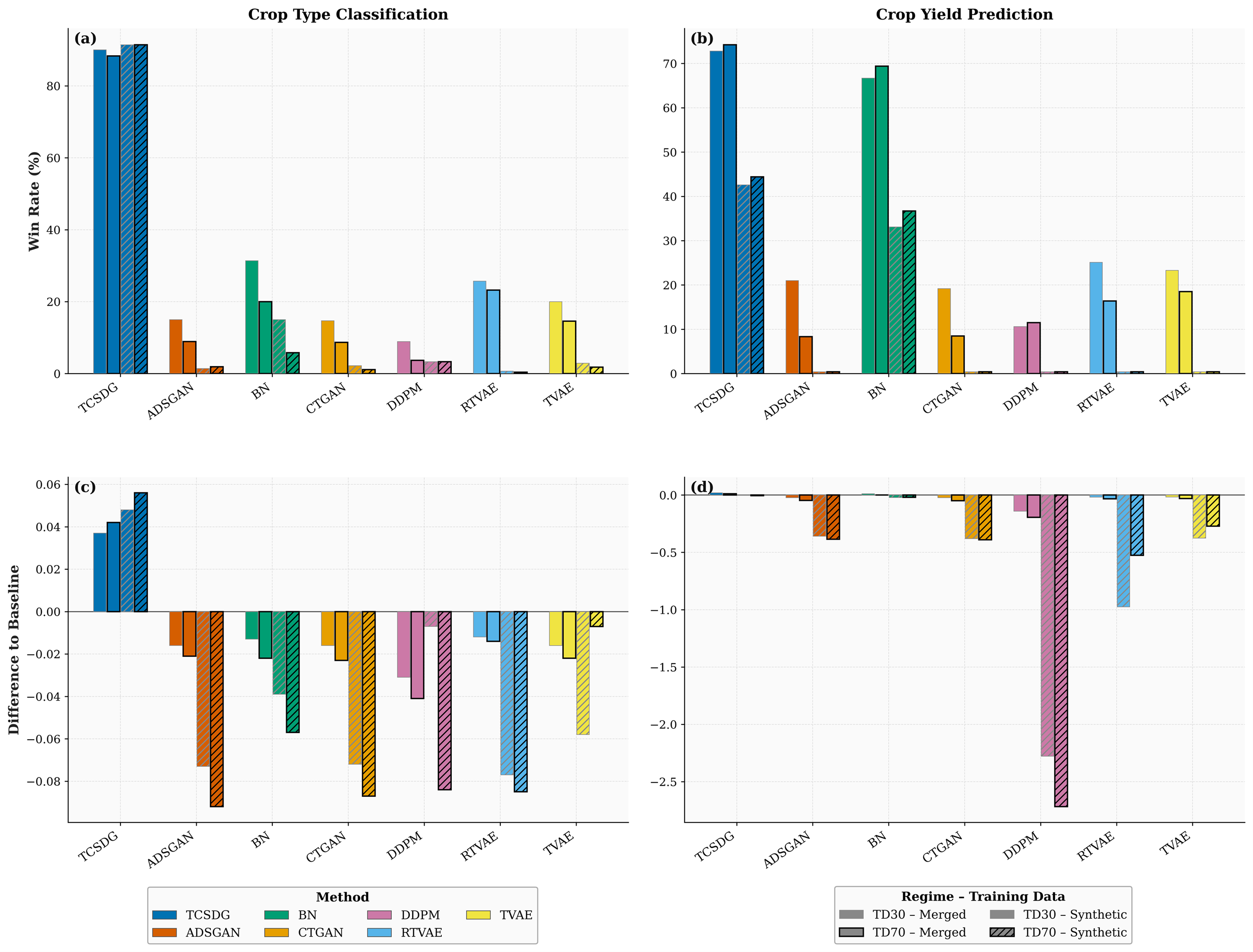}
    \caption{Effect of the real-data-training fraction (TD30 and TD70) on utility of SDG methods, under merged and synthetic regimes. (a, b) Win rate (\%) over the no-augmentation baseline for crop type classification and crop yield prediction. (c, d) Mean signed difference to the baseline in the metric's natural units (BAC for classification, sign-flipped RMSE for regression).}
    \label{fig:DataSizeImapct}
\end{figure*}

For each SDG method and evaluation regime, the win rate against the real-data baseline and the mean difference relative to that baseline at two real-training-data fractions (TD30 and TD70) is presented in Figure ~\ref{fig:DataSizeImapct}. In crop-type classification, TCSDG remains consistently superior under both data fractions. In the merged regime, it achieves win rates of 90\% at TD30 and 88\% at TD70, while in the synthetic-only regime its win rate remains stable at 91\% under both conditions. By contrast, all benchmark SDG methods remains below the baseline in crop type classification at both training sizes. For crop-yield prediction, the effects are smaller but still favor TCSDG. In the merged regime, TCSDG achieves win rates of 73\% at TD30 and 74\% at TD70. BN was the only other generator that shows systematic positive utility for crop yield prediction, with its win rate increasing from 67\% to 69\%. The remaining SDG methods show low win rates, and their synthetic-only results are effectively unusable.

For most benchmark SDG methods, increasing the amount of real-training-data weakened their relative position because the real-only baseline became stronger and therefore more difficult to outperform. Under the merged regime, classification win rates decline across the benchmark methods, including BN from 31\% to 20\%, TVAE from 20\% to 15\%, and DDPM from 9\% to 4\%. A similar pattern was observed in crop yield prediction, where the GAN-based generators lost much of their relative utility; for example, ADSGAN declined from 21\% to 8\% and CTGAN from 19\% to 9\%. In the synthetic regime, DDPM deteriorates further, with its mean gap from the baseline widening from $-$2 at TD30 to $-$3 at TD70. The only partial exception among the benchmark SDG methods was BN in yield prediction under the merged regime, where it remains above the baseline and improved slightly as more real-training-data became available. TCSDG showed the opposite pattern as rather than losing utility as the real-training-data fraction increased, it maintained or extended its advantage over the baseline.

This pattern was notable because many tabular SDG methods are motivated by limited-data settings and are commonly evaluated under data-scarcity assumptions \citep{akkem_comprehensive_2024,fonseca_tabular_2023,jamnik_tabebm_2024}. Previous work has also shown that, for some SDG approaches, the largest performance gains occur in the smallest-data regime and diminish as more real data become available \citep{ebrahimy_utilization_2023}. TCSDG departed from this scarcity-driven pattern, since its utility persisted, and in classification even increased, when the real-training-data fraction rose from 30\% to 70\%. Methodologically, this persistence suggested that TCSDG captured information that was not redundant with the real training data; otherwise, its marginal benefit would have been expected to shrink toward zero as the real-data baseline strengthened.

\subsection{Per-dataset breakdown}
\label{sec:3.5}
At the dataset level with fixed multiplication ratio of 2 and merged regime (see Table ~\ref{tab:per_dataset_impact}), TCSDG shows the most stable performance among all generators. For the crop type classification task, TCSDG attained the highest mean BAC across all six datasets and was the only generator to produce statistically significant improvements over the baseline in all the experiments. The remaining SDG methods (ADSGAN, BN, CTGAN, DDPM, RTVAE, and TVAE) consistently fail to surpass the baseline and, in several cases most notably DDPM in Germany and Romania, aggressively degrade downstream classification quality.

\begin{table*}[!h]
\centering
\caption{The per-dataset breakdown of SDG performance at a fixed multiplication ratio of 2 and merged regime, with each cell summarizing 60 paired evaluations (10 repetitions × 3 downstream ML learners × 2 real training-data-fractions). The values represent mean BAC for crop type classification and RMSE for crop yield prediction with standard deviation values provided in parenthesis. The bold values show the most accurate method in each dataset, and the * sign marks entries that outperform baseline on a paired Wilcoxon signed-rank test, with the Holm correction applied within each row across the seven SDG method-vs-baseline hypotheses.}
\label{tab:per_dataset_impact}
\footnotesize
\setlength{\tabcolsep}{4pt}
\begin{tabular}{l l c ccccccc}
\toprule
Task & Dataset & Baseline & TCSDG & ADSGAN & BN & CTGAN & DDPM & RTVAE & TVAE \\
\midrule
\multirow{6}{2cm}{Crop Type Classification}
 & Austria  & 0.23(0.07) & \textbf{0.28(0.04)*} & 0.21(0.06) & 0.21(0.04) & 0.21(0.05) & 0.19(0.06) & 0.22(0.05) & 0.21(0.04) \\
 & Bulgaria & 0.17(0.05) & \textbf{0.22(0.03)*} & 0.16(0.04) & 0.17(0.04) & 0.16(0.04) & 0.16(0.05) & 0.16(0.04) & 0.16(0.04) \\
 & Germany  & 0.12(0.03) & \textbf{0.15(0.02)*} & 0.10(0.02) & 0.10(0.02) & 0.10(0.02) & 0.09(0.01) & 0.11(0.02) & 0.10(0.02) \\
 & Greece   & 0.22(0.04) & \textbf{0.25(0.03)*} & 0.21(0.03) & 0.21(0.03) & 0.21(0.03) & 0.18(0.06) & 0.21(0.03) & 0.20(0.03) \\
 & Italy    & 0.18(0.04) & \textbf{0.21(0.02)*} & 0.17(0.03) & 0.16(0.02) & 0.17(0.03) & 0.16(0.03) & 0.17(0.03) & 0.15(0.03) \\
 & Romania  & 0.13(0.04) & \textbf{0.17(0.03)*} & 0.12(0.03) & 0.12(0.03) & 0.12(0.02) & 0.10(0.02) & 0.13(0.03) & 0.12(0.03) \\
\midrule
\multirow{6}{2cm}{Crop Yield Prediction}
 & Germany   & 1.09(0.10) & \textbf{1.07(0.07)*} & 1.10(0.06) & 1.09(0.09)* & 1.10(0.06) & 1.27(0.19) & 1.09(0.06)* & 1.09(0.06)* \\
 & India     & 0.54(0.03) & \textbf{0.54(0.04)}  & 0.58(0.05) & 0.54(0.05)  & 0.58(0.05) & 0.64(0.09) & 0.57(0.07)  & 0.57(0.06)  \\
 & Zambia    & \textbf{0.58(0.05)} & 0.58(0.06)  & 0.59(0.04) & 0.59(0.07)  & 0.60(0.05) & 0.67(0.10) & 0.60(0.06)  & 0.60(0.06)  \\
 & Argentina & 1.36(0.13) & \textbf{1.35(0.13)*} & 1.42(0.11) & 1.36(0.14)  & 1.44(0.11) & 1.64(0.29) & 1.41(0.13)  & 1.41(0.11)  \\
 & France    & 0.78(0.14) & \textbf{0.72(0.10)*} & 0.79(0.10) & 0.73(0.10)* & 0.79(0.10) & 0.94(0.14) & 0.78(0.09)  & 0.77(0.09)  \\
 & Italy     & 0.82(0.11) & \textbf{0.80(0.11)*} & 0.83(0.09) & 0.80(0.11)* & 0.83(0.10) & 0.88(0.10) & 0.83(0.11)  & 0.84(0.10)  \\
\bottomrule
\end{tabular}
\end{table*}

The crop yield prediction results were more nuanced but remained consistent with the classification findings. TCSDG achieved the best mean RMSE on five of six datasets and significantly improved performance on Germany, Argentina, France, and Italy. On India, TCSDG matches the baseline (both 0.54) with no meaningful difference, while Zambia was the only dataset where the baseline retains the lowest (best) mean RMSE. Even in Zambia, the TCSDG–baseline difference was within one standard deviation and did not indicate a significant loss. BN was again the only credible non-TCSDG alternative, producing significant RMSE reductions on Germany, France, and Italy. DDPM was consistently the weakest method, increasing RMSE values on every dataset.

The robustness of TCSDG across country-level LUCAS crop classification tasks and CY-Bench yield-prediction tasks suggests that its advantage was not caused by a single favorable dataset. In both tasks the relative position of each generator was essentially identical from dataset to dataset, supporting the view that the observed effects reflect genuine differences in generator quality rather than artefacts of any particular dataset or experimental configuration. The cross-dataset stability observed here therefore strengthens the case for TCSDG as a task-aware augmentation method.

\subsection{Implications and Future Direction}
\label{sec:3.6}
The results carry three methodological implications for synthetic data in agriculture. First, the utility of a tabular SDG method is determined by whether it preserves the conditional structure required by the downstream task, not by how realistically it reproduces the joint distribution; methods optimized for fidelity alone do not consistently transfer to the agricultural settings examined here. Second, the synthetic multiplication ratio is not a neutral scaling parameter: when the generator is task-misaligned, larger synthetic budgets actively erode downstream performance, whereas a task-conditioned generator such as TCSDG exhibits controlled saturation. Third, generator choice cannot be decoupled from the choice of downstream learner: learners that support sample weighting realize the largest gains from TCSDG, while unweighted learners respond more conservatively. Taken together, these observations argue for a unified, task-conditioned design and evaluation pipeline for agricultural synthetic data, in which the generator, the candidate-selection stage, and the downstream training procedure are co-specified rather than treated as independent modules. 

Several scope decisions in this work define its natural extensions. The benchmark is utility-centered, consistent with the ML-efficiency evaluation paradigm of the tabular SDG literature; fidelity and privacy diagnostics, for example, the distance-to-closest-record analysis used to detect memorization of training records \citep{borisov_language_2023}, are complementary directions that can be layered onto the present framework. Moreover, a promising direction is to leverage tabular foundation models (e.g., TabICL) to design a model-agnostic framework that operates on top of arbitrary SDG methods and prunes their generated samples into task-aware subsets tailored to the intended downstream application. Such a framework would decouple the synthetic data quality from the choice of generator and could provide a principled mechanism for selecting samples that maximize downstream utility rather than purely statistical fidelity.

\section{Conclusion}
This study introduced TCSDG, a task-conditioned synthetic data generation algorithm that combines class- or bin-aware budget allocation, a teacher-based consistency filter, coverage-preserving selection, and quality-aware sample weighting into a single pipeline. The algorithm was benchmarked against six established tabular SDG methods on twelve supervised agricultural tasks under identical training regimes, multiplication ratios, and downstream ML learners. TCSDG was the only method in the benchmark to deliver statistically reliable improvements over the no-augmentation baseline in both tasks under the merged regime, and for crop type classification under the synthetic regime. For crop yield prediction under the synthetic regime, TCSDG remained the best-performing method, although its mean improvement was not statistically significant. Its advantage held across multiplication ratios, with performance saturating rather than collapsing as the synthetic fraction grew; across downstream learners, with the largest gains for RF and SVM that exploit the teacher-derived instance weights; and across all twelve country-level datasets. Crucially, the benefit did not vanish when real training data became more abundant, indicating that TCSDG supplies task-relevant variation that is complementary, rather than redundant, to the real training set.

\section{Acknowledgements}
This work was supported by the Lower Saxony Ministry of Science and Culture (MWK), funded through the zukunft.niedersachsen program of the Volkswagen Foundation (ZN4072).

\section{Data availability statement}
The data utilized in this study are openly available. Specifically, European Land Use/Cover Area frame Survey (LUCAS) dataset 2022 is available at \href{https://ec.europa.eu/eurostat/web/lucas/database/2022}{https://ec.europa.eu/eurostat/web/lucas/database/2022}, while the version 1.4 of CY-Bench dataset is available at \href{https://doi.org/10.5281/zenodo.14243115}{https://doi.org/10.5281/zenodo.14243115}.

\bibliographystyle{elsarticle-harv}
\bibliography{TCSDG_references.bib}

\end{document}